\documentclass[sigconf,natbib=false]{acmart}
\usepackage{caption}
\usepackage{subcaption}
\usepackage[style=ACM-Reference-Format,maxbibnames=4,maxcitenames=1, backend=biber,uniquelist=false,natbib,doi=false,isbn=false,url=false,giveninits]{biblatex}
\renewbibmacro{in:}{}

\usepackage{graphicx}
\usepackage{tikz}
\usetikzlibrary{fit}
\usepackage{mathtools}
\usepackage[boxed]{algorithm2e}
\usepackage{xcolor}
\usepackage[T1]{fontenc}
\usepackage[utf8]{inputenc}

\definecolor{green1}{RGB}{0,153,0}
\definecolor{red1}{RGB}{204,0,0}
\definecolor{blue1}{RGB}{0,0,153}
\usepackage{color}
\usepackage{enumitem}

\newcommand{\ourmethod}{Grale}

\newcommand{\videosite}{YouTube}
\newcommand{\bigO}[1]{\mathcal{O}(#1)}
\usepackage{combelow}

\makeatletter
\g@addto@macro\normalsize{%
  \setlength\abovedisplayskip{2pt}
  \setlength\belowdisplayskip{2pt}
  \setlength\abovedisplayshortskip{2pt}
  \setlength\belowdisplayshortskip{2pt}
}
\makeatother

\title{Grale: Designing Networks for Graph Learning}
\author{Jonathan Halcrow$^\dagger$, Alexandru Mo\c{s}oi$^\ddagger$, Sam Ruth$^\dagger$, Bryan Perozzi$^\dagger$}
\affiliation{$\dagger$: Google Research}
\affiliation{$\ddagger$:  YouTube}
\email{{halcrow, mosoi, samruth}@google.com, bperozzi@acm.org}
\copyrightyear{2020}
\acmYear{2020}
\setcopyright{rightsretained}
\acmConference[KDD '20]{Proceedings of the 26th ACM SIGKDD
Conference on Knowledge Discovery and Data Mining}{August 23--27,
2020}{Virtual Event, CA, USA}
\acmBooktitle{Proceedings of the 26th ACM SIGKDD Conference on
Knowledge Discovery and Data Mining (KDD '20), August 23--27, 2020,
Virtual Event, CA, USA}
\acmDOI{10.1145/3394486.3403302}
\acmISBN{978-1-4503-7998-4/20/08}

\begin{abstract}
How can we find the right graph for semi-supervised learning? 
In real world applications, the choice of which edges to use for computation is the first step in any graph learning process.
Interestingly, there are often many types of similarity available to choose as the edges between nodes, and the choice of edges can drastically 
affect the performance of downstream semi-supervised learning systems.
However, despite the importance of graph design, most of the literature assumes that the graph is static.

In this work, we present \ourmethod, a scalable method we have developed to address the problem of graph design for graphs with billions of nodes.
\ourmethod\ operates by fusing together different measures of (potentially weak) similarity to create a graph which exhibits high task-specific homophily between its nodes.
\ourmethod\ is designed for running on large datasets. We have deployed \ourmethod\ in more than 20 different industrial settings at Google, including datasets which have  \emph{tens of billions} of nodes, and \emph{hundreds of trillions} of potential edges to score. By employing locality sensitive hashing techniques, we greatly reduce
the number of pairs that need to be scored, allowing us to learn a task specific model and build the associated nearest neighbor graph for such datasets in hours, rather than
the days or even weeks that might be required otherwise.

We illustrate this through a case study where we examine the application of \ourmethod\ to
an abuse classification problem on \videosite\ with hundreds of million of items.
In this application, we find that \ourmethod\ detects a large number of
malicious actors on top of hard-coded rules and content classifiers,
increasing the total recall by 89\% over those approaches alone.
\end{abstract}

\begin{document}

\maketitle

\renewcommand{\shortauthors}{Halcrow et al.}

\newpage
\section{Introduction}

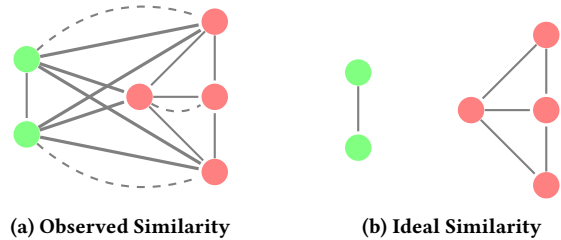
\begin{figure}[t!]
\vspace{-0.2cm}
\centering
	\centering	
	\def\gralelayersep{2cm}
	\begin{subfigure}[t]{0.48\columnwidth}
	\centering	
	\begin{tikzpicture}[shorten >=1pt,draw=black!50, node distance=\gralelayersep]
		\tikzstyle{every pin edge}=[<-,shorten <=1pt]
		\tikzstyle{neuron}=[circle,fill=black!25,minimum size=10pt,inner sep=0pt]
		\tikzstyle{pw neuron}=[neuron, fill=black!50];
		\tikzstyle{good node}=[neuron, fill=green!50];
		\tikzstyle{bad node}=[neuron, fill=red!50];
		\tikzstyle{annot} = [text width=4em, text centered]
		
		\tikzstyle{similarity1}=[very thick, fill=black!100];
		\tikzstyle{similarity2}=[thick];
		\tikzstyle{similarity3}=[thick, dashed];		
		
		\node[good node] (I-1) at (0.5,-1) {};
		\node[good node] (I-2) at (0.5,-2) {};	

		\node[bad node] (H-1) at (3cm,-0.5 cm) {};
		\node[bad node] (H-2) at (\gralelayersep,-1.5 cm) {};
		\node[bad node] (H-3) at (3cm,-2.5 cm) {};
		
		\node[bad node] (O) at (3cm,-1.5 cm) {};
		
		\foreach \source in {1,...,2}
		\foreach \dest in {1,...,3}
		\draw[similarity1] (I-\source) edge (H-\dest);
		
		\draw[similarity3, bend left] (I-1) edge (H-1) {};
		\draw[similarity3, bend right] (I-2) edge (H-3) {};						
		\draw[similarity3, bend right] (H-2) edge (O);						
		
		\foreach \source in {1,...,3}
		\draw[similarity2] (H-\source) edge (O); 
		
		\draw[similarity2] (H-1) edge (H-2);
		\draw[similarity2] (H-2) edge (H-3);	
		\draw[similarity2] (I-1) edge (I-2);

		\end{tikzpicture}
		\caption{Observed Similarity}
	\end{subfigure}
 	\hfill
	\begin{subfigure}[t]{0.48\columnwidth}
	\centering			
	\begin{tikzpicture}[shorten >=1pt,draw=black!50, node distance=\gralelayersep]
	\tikzstyle{every pin edge}=[<-,shorten <=1pt]
	\tikzstyle{neuron}=[circle,fill=black!25,minimum size=10pt,inner sep=0pt]
	\tikzstyle{pw neuron}=[neuron, fill=black!50];
	\tikzstyle{good node}=[neuron, fill=green!50];
	\tikzstyle{bad node}=[neuron, fill=red!50];
	\tikzstyle{annot} = [text width=4em, text centered]
	
	\tikzstyle{similarity1}=[very thick, fill=black!100];
	\tikzstyle{similarity2}=[thick];
	
	\node[good node] (I-1) at (0.5,-1) {};
	\node[good node] (I-2) at (0.5,-2) {};	

	\node[bad node] (H-1) at (3cm,-0.5 cm) {};
	\node[bad node] (H-2) at (\gralelayersep,-1.5 cm) {};
	\node[bad node] (H-3) at (3cm,-2.5 cm) {};
	
	\node[bad node] (O) at (3cm,-1.5 cm) {};
	
	\foreach \source in {1,...,3}
	\draw[similarity2] (H-\source) edge (O); 
	
	\draw[similarity2] (H-1) edge (H-2);
	\draw[similarity2] (H-2) edge (H-3);	
	\draw[similarity2] (I-1) edge (I-2);	
	\end{tikzpicture}
	\caption{Ideal Similarity}
	\end{subfigure}	
\caption{The Graph Design Problem.  On the left is an example of a typical product graph with several different kinds of possible similarity (edge types) connecting nodes of two types (green,red).  
While informative, this graph can be challenging for semi-supervised learning methods -- even though the products share some facets of similarity, they are not informative for the final task.
On the right, is the ideal graph similarity which would maximize label classification performance on this dataset. 
In this work, we present \ourmethod, a scalable method for extracting graphs from very large datasets.}
\label{fig:crownjewel}
\end{figure}

As the size and scope of unlabeled data has grown, there has been a corresponding surge of interest in graph-based methods for Semi-Supervised Learning (SSL) (e.g.\cite{blum2001learning}\cite{LGCpaper}\cite{yang2016revisiting}\cite{chami2020machine}) which make use of both the labeled data items and also of the structure present across the whole dataset.
An essential component of such SSL algorithms is the \emph{manifold assumption} \textemdash the assumption that the labels vary smoothly over the underlying graph structure (i.e. that the graph possesses homophily).
As such, the datasets that are typically used to evaluate these methods are small datasets that exhibit this property (e.g. citation networks).
However, there is relatively little guidance on how to apply these methods when there is not an obvious source of similarity that's useful for the problem \cite{leman2018cikm}.

Real-world applications tend to be multi-modal, and so this lack of guidance is especially problematic.
Rather than there being just 
one simple feature space to choose a similarity on, we have a wealth of different modes to compare
similarities in (each of which may be suboptimal in some way \cite{deSousa2013}).
For example, video data comes with visual signals, audio signals, text signals,
and other kinds of metadata - each with their own ways of measuring similarity. 
Through work on many applications at Google we have found that the right answer is 
to choose a combination of these.
But how should we choose this combination? 

In this paper we present \ourmethod, a method we have developed to design graphs for SSL.
Our work on \ourmethod\ was motivated by practical problems faced while applying SSL to rich heterogeneous datasets in an industrial setting.
\ourmethod\ is capable of learning similarity models and building graph
for datasets with \emph{billions} of nodes in a matter of hours. It
is widely used inside Google, with more than 20 deployments,
where it acts as a critical component of semi-supervised learning systems.
At its core, \ourmethod\ provides the answer to a very straightforward question -- given a semi-supervised learning task -- 
``do we want this edge in the graph?''.

We illustrate the benefits of \ourmethod\ with a case study of a real anti-abuse detection system at YouTube, a large video sharing site.
Fighting abuse on the web can be a challenging problem area because it falls into an area
where features that one might have about abusive actors are quite sparse compared to the number
of labeled examples available at training time. 
Abuse on the web is also often done at
scale, inevitably leaving some trace community structure that can be exploited to fight it.

Standard supervised approaches often fail in this type of setting since there are not enough labeled
examples to adequately fit a model with capacity large enough to capture the complexity of the problem.

Specifically, our contributions are the following:
\begin{enumerate}[leftmargin=0.5cm,itemindent=.5cm,labelwidth=\itemindent,labelsep=0cm,align=left,topsep=0pt,itemsep=-1ex,partopsep=1ex,parsep=1ex]
    \item An algorithm and model for learning a task specific graph for semi-supervised applications.
    \item An efficient algorithm for building a nearest neighbor graph with such a model.
    \item A demonstration of the effectiveness of this approach for fighting abuse on \videosite, with hundreds of millions of items.
\end{enumerate}

\section{Preliminaries}

\subsection{Notation}
\label{subsec:setting}
We consider a general multiclass learning setting where we have a partially labeled set of points
$\mathcal{X} = \{x_1, x_2, \dots, x_V\}$ where the first $L$ points have
class labels $\mathbf{Y} = \{\mathbf{y}_1, \mathbf{y}_2, \dots, \mathbf{y}_L\}$, each $\mathbf{y}_k$ being a
one-hot vector of dimension $C$, indicating which of the $C$ classes the associated point belongs to. 
Further, we assume each point $x_i \in \mathcal{X}$ has a multi-modal representation over $d$ modalities,
$x_i = \{{x}_{i,1}, {x}_{i,2}, \dots {x}_{i, d}\}$. 
Each sub-representation $x_{i,d}$ has its own natural 
distance measure $\kappa_d$.

\subsection{Loss functions}
Ordinarily one might try to fit a model to make predictions of $y_k = \hat{y}(\mathbf{x_k})$
by selecting a family of functions and finding a member of it which minimizes the
cross-entropy loss between $y$ and $\hat{y}$:
\begin{equation}
    \mathcal{L} = -\sum_{i\in \mathcal{Y}}\sum_{c \in \mathcal{C}} y_{i,c} \log{\hat{y}_{i, c}} ,
    \label{equ:logloss}
\end{equation}
where $y_{i,c}$ and $\hat{y}_{i, c}$ indicate the ground truth and prediction scores
for point $i$ with respect to class $c$. 

Given some graph $G=(V,E)$ with edge weights $w_{ij}$ on our data, SSL graph algorithms \cite{ravi2016large,LGCpaper,murua2008potts} nearly universally seek to minimize a 
Potts model type loss function (either explicitly or implicitly) similar to:
\begin{equation}
\label{ising_loss}
\mathcal{L} = \sum_{i,j \in E} w_{i,j} \sum_{c \in \mathcal{C}} |\hat{y}_{i, c} - \hat{y}_{j, c}| + \sum_{i \in V, c \in \mathcal{C}} |\hat{y}_{i,c} - y_{i,c}| .
\end{equation}

A classic way to minimize this type of loss is to iteratively apply a label update rule of the form:
\begin{equation}
    \label{equ:regular_update}
    \hat{y}^{(n+1)}_{i,c} = \alpha y_{i,c} + \beta \frac{\sum_{j\in \mathcal{N}_i} w_{i,j} \hat{y}^{(n)}_{j,c}}{\sum_{j \in \mathcal{N}_i} w_{i,j}}  ,
\end{equation}
where $\mathcal{N}_i$ is the neighborhood of point $i$, $\hat{y}^{(n)}_{i,c}$ is the predicted
score for point $i$ with respect to class $c$ after $n$ iterations, and $\alpha$ and $\beta$ are hyperparameters.
Here, we consider a greedy solution which seeks to minimize Eq.~\eqref{equ:logloss}
using a single iteration of label updates.

\subsection{The Graph Design Problem}

In Equations~\eqref{ising_loss} and~\eqref{equ:regular_update} it is assumed the edge weights $w_{i,j}$ are given to us. 
However in the multi-modal setting we describe here it is not the case that there is a single obvious choice. 
This is a critical decision
when building such a system, which can strongly impact the overall accuracy \cite{deSousa2013}.
In many cases one might heuristically choose some similarity measure, performing 
some hyperparameter optimization routine to select the number of neighbors to compare to
or some $\epsilon$ similarity threshold. When one includes ways of selecting or combining various
similarity measures, the parameter space can become too large to feasibly handle with a simple grid search
- especially when the dataset being operated on becomes large.

Instead we tackle this problem head on, framing the problem as one of graph design rather than 
graph selection. The Graph Design problem is as follows:\\
\textbf{Given}:
\begin{itemize}[leftmargin=0.25cm,itemindent=.25cm,labelwidth=\itemindent,labelsep=0cm,align=left,topsep=0pt,itemsep=-1ex,partopsep=1ex,parsep=1ex]
    \item A multi-modal feature space $\mathcal{X}$ (as in Subsection~\ref{subsec:setting})
    \item A partial labeling on this feature space $\mathcal{Y}$
    \item A learning algorithm $\mathcal{A}$ which is a function of some graph $G$ having
          vertex set equal to the elements of $\mathcal{X}$
\end{itemize}
\textbf{Find}:
An edge weighting function $w(x_i, x_j)$ that allows us to construct a graph $G$ which optimizes
the performance of $\mathcal{A}$.

\section{\ourmethod}

\ourmethod\ is built to efficiently solve this type of problem - designing task specific graphs by reducing
the original task to training a classifier on pairs of points.  We assume the existence of some
oracle $y(x_i, x_j)$ which gives a binary valued ground truth, defined for some subset of unordered pairs $(x_i, x_j)$.  The oracle is chosen to produce a graph amenable to
the primary task, $\mathcal{A}$; in the multiclass learning setting described above, the relevant oracle would
return true if and only if both points are labeled and share the same class.

\subsection{A Pairwise Loss for Graph Design}

We start with the nearest neighbor update rule from  Eq.~\eqref{equ:regular_update} and let the weights be the $\log$ of some function
of our `natural' metrics in each mode: $w_{ij} := \log G(x_i, x_j)$, for some $G: \mathbf{R}^d x \mathbf{R}^d \rightarrow \mathbf{R}$, which only depends on the $d$ modality distances:
\begin{equation}
     G(x_i, x_j) = f(\kappa_1(x_{i}, x_{j}), \kappa_2(x_{i}, x_{j}), \dots, \kappa_d(x_{i}, x_{j})).
\end{equation}

Next, we choose a relaxed form of Eq.~\eqref{equ:regular_update}, where we drop the normalization
factor and instead impose a constraint $w_{ij} < 0$
\begin{equation}
    \label{equ:our_update}
    \hat{y}_{i,c} =  \prod_{j \in \mathcal{N}_{i,c}} G(x_i, x_j) \,
\end{equation}
with $\mathcal{N}_{i,c}$ being the set of neighbors of node $i$ with $y_{j,c} = 1$.

Removing the normalization factor greatly simplifies the optimization problem.
Applied to $G$, our new constraint is transformed to requiring that $0 < G < 1$. This can
be achieved simply by choosing $G$ to belong to some bounded family of functions (such as applying
a sigmoid transform to the output of a neural network). Note also that since
each $\kappa$ is a metric and thus symmetric in its arguments this also gives us that $w_{ij} = w_{ji}$, ensuring
that a nearest neighbor graph we build with this edge weighting function will be undirected.
Putting this back into the full multi-class cross-entropy loss on the node labels Eq.~\eqref{equ:logloss}, we recover the loss of a somewhat simpler problem
to solve: the binary classification problem of deciding whether two nodes are both members
of the same class.
\begin{equation}
    \mathcal{L} = -\sum_{c \in \mathcal{C}} \sum_{i \in \mathcal{X}} \sum_{j \in \mathcal{X}} y_{i,c}, y_{j,c} \log G(x_i, x_j).
\end{equation}

\subsection{Locality Sensitive Hashing}
\label{subsec:lsh}
A key requirement for \ourmethod\ is that it must scale to datasets containing billions of nodes, making an all-pairs
search infeasible. Instead we rely on approximate similarity search using locality sensitive 
hashing (LSH). Generally speaking LSH works by selecting a family of hash functions $\mathcal{H}$ with the property
that two points which hash to the same value are likely to be `close'. Then one compares
all such points which share a hash value. In our case,
however, the notion of similarity is the output of a model. So we don't have an obvious choice
of LSH function. 
Here we analyze two different ways to construct such a LSH function for our
input distances: AND-amplification and OR-amplification.

Recall that above we assume $G(\mathbf{x}_i, \mathbf{x}_j)$ is purely a function of distances
between $\mathbf{x}_i$ and $\mathbf{x}_j$ in $D$ subspaces. Let us define a metric $\mu$ on
$\mathcal{X} \times \mathcal{X}$ as a sum over our individual metrics $\kappa_d$:
\begin{equation}
    \mu\big((x_1, x_2), (x_3, x_4)\big) = \sum_d \kappa_d(x_1, x_3) + \kappa_d(x_2, x_4) .
\end{equation}
Let us also assume that our similarity model $G$ is Lipschitz continuous on 
$\mathcal{X} \times \mathcal{X}$, namely that there exists some $K \in \mathbf{R}$ such that:
\begin{equation}
    |G(x_i, x_j) - G(x_k, x_l)| \leq K \mu\big((x_i, x_j), (x_k, x_l)\big) .
\end{equation}
This implies that
\begin{equation}
    |G(x_i, x_i) - G(x_i, x_j)| \leq K \sum_{d} \kappa_d(x_i, x_j) .
    \label{eq:gdiff}
\end{equation}

Next, let $\Delta(x_i, x_j) = 1- G(x_i, x_j)$.
We note that $\Delta(x_i, x_j)$
is not precisely a metric, but it is bounded to the unit interval. Applying this to Eq.\ \eqref{eq:gdiff} yields:
\begin{equation}
    |\Delta(x_i, x_i) - \Delta(x_i, x_j)| \leq K \sum_{d} \kappa_d(x_i, x_j) .
\end{equation}

In practice $\Delta(x_i, x_i)$ is quite small (a point always has the same label as itself!), but
a learned model may not evaluate to exactly 0. Let $\epsilon = \sup_{x \in \mathcal{X}} \Delta(x, x)$, 
either $\Delta(x_i, x_j) < \epsilon$ or we have
\begin{equation}
    \Delta(x_i, x_j) - \epsilon \leq  \Delta(x_i, x_j) - \Delta(x_i, x_i) \leq K \sum_{d} \kappa_d(x_i, x_j) .
\end{equation}
\noindent In either case,
\begin{equation}
\vspace{-0.2cm}
      \Delta(x_i, x_j) \leq K \sum_{d} \kappa_d(x_i,  x_j) + \epsilon .
\end{equation}
\par
\medskip
Next let $\mathcal{H}_n$ be an
$(r, cr, p, q)$ sensitive family of locality sensitive hash functions for $\kappa_n$. That is,
two points $x_i$ and $x_j$ with $\kappa_n(x_i, x_j) \leq r$ have probability at least $p$
of sharing the same hash value for some hash function $h$ randomly chosen from $\mathcal{H}_n$. 
Further two points $x_i$ and $x_j$ with $\kappa_n(x_i, x_j) \geq cr$ have probability of at most $q$
of sharing the same hash value for a similarly chosen $h$.

We construct a new `AND' family $\mathcal{H}_\otimes$ of hash functions by concatenating hash functions 
from each of the $\mathcal{H}_n$, with each being $(r, c_n r, p_n, q_n)$ sensitive for the
associated space. Then $\mathcal{H}_\otimes$ is $(r_\otimes, cr_\otimes, p_\otimes, q_\otimes)$-sensitive for
$G$, where 
\begin{equation}
\begin{aligned}
    r_\otimes &= KrD + D \epsilon \\
    p_\otimes &= \prod_{d} p_d \\
    q_\otimes &= \prod_{d} q_d .
\end{aligned}
\end{equation}

Alternatively we may also employ an `OR'-style construction of an LSH function, taking $\mathcal{H}_\oplus$ as
the union of the $\mathcal{H}_n$. If we assume that there exists some correlation between the subspaces of
$\mathcal{X}$ such that two points having $\kappa_m(x_i, x_j) \leq r$ also have $\kappa_n(x_i, x_j) \leq r$
with probability $p_{mn}$ (and likewise for distance $\geq cr$ with probability $q_{mn}$) then 
$\mathcal{H}_m$ acts as a locality sensitive hash function for $\mathcal{H}_n$.
Under this assumption $\mathcal{H}_\oplus$ is $(r_\oplus, cr_\oplus, p_\oplus, q_\oplus)$-sensitive for $G$,
where
\begin{equation}
\begin{aligned}
    r_\oplus &= KrD + D \epsilon \\
    p_\oplus &= \min_d \left(1 - (1-p_d )\prod_{l \neq d} (1 - p_{dl}) \right) \\
    q_\oplus &= \max_d \left(1 - (1-q_d) \prod_{l \neq d} (1 - q_{dl}) \right).
\end{aligned}
\end{equation}

In practice, we employ a mixture of the AND and OR forms of locality sensitive hash functions when applying
\ourmethod, tuning choices of functions and hyper-parameters by grid search (see Table~\ref{tab:sketch_perf}
for an evaluation of the performance on a real world dataset). 

\subsection{Graph building algorithm}

\begin{algorithm}[t]
\vspace{-0.1cm}
\caption{NNSketching($\mathcal{X}$)}
\SetKwFunction{NNSketching}{NNSketching}
\SetKwFunction{LSH}{LSH}
\SetKwData{B}{Buckets}
\SetKwData{K}{K}
\SetKwData{sketches}{sketches}
\SetKwData{bucket}{bucket}
\SetKwData{weightij}{Gij}
\SetKwFunction{GraleModel}{y}
\SetKwData{Graph}{G}
\SetKwData{threshold}{$\epsilon$}
\SetKwProg{Fn}{Function}{:}{}
\SetKwInOut{Input}{Input}\SetKwInOut{Output}{Output}

\Fn{\NNSketching{$\mathcal{X}$}}{
\Input{A set of points $\mathcal{X}$}
\ForEach{$p \in \mathcal{X}$}{
\ForEach{$h \in \LSH{p}$}{
Append $h$, $p$ to \sketches\;
}
}
Sort \sketches by hash value \;
Collect \sketches into bucket with same hash $\rightarrow$ \B\;
Subdivide buckets in \B of size larger than \K\;
\Return{\B}\;
}

\BlankLine

\SetKwFunction{BuildGraph}{BuildGraph}
\Fn{\BuildGraph{$\mathcal{X}$, $\hat{y}$, $\epsilon$}}{
\Input{A set of points $\mathcal{X}$, similarity model $\hat{y}$, and minimum edge weight \threshold}
\Output{A nearest neighbor graph \Graph}
\BlankLine
\B = \NNSketching{$\mathcal{X}$}
 \BlankLine
 \ForEach{$\bucket \in \B$} {
     \ForEach{pair $p_i, p_j \in \bucket$} {
        $w_{ij} = \hat{y}(p_i, p_j)$ \;
         \If{$w_{ij} > \threshold$}{
             Emit edge $(p_i, p_j, w_{ij})$ \;
         }
     }
}
\Return{$\Graph = \cup$ edges}
}
\caption{Graph Building Algorithm}
\label{alg:graphbuilder}
\end{algorithm}

Our graph building algorithm works by employing the LSH function constructed for our similarity measure to produce
candidate pairs, then scoring the candidate pairs with the model.
More explicitly, we start by sampling $S$ hash functions from our chosen family $\mathcal{H}_s$ (constructed
from the set of LSH families for our features). Then for each point $p_i \in X$ we compute a set of
hashes, bucketing by hash value. Finally, we find it practical to cap the size of each bucket (typically this is 
chosen to be 100), randomly subdividing any bucket which is larger.
Without this step in the worse case we could still be computing all $O(N^2)$ comparisons. An additional practical 
step we take in some cases is to simply drop buckets which are too large, since this is usually indicative of 
hashing on a 'noise' feature and this buckets tend not to contain many desirable pairs, and often lead
to false positives from the model.
For each pair in our subdivided buckets we compare all pairs with our scoring function $G$ (learned using
the training procedure described in the next section). Finally we output the top $k$ values (potentially
subject to some $\epsilon$ threshold on the scores). 

\subsection*{Training algorithm for Grale}
During the learning phase, we follow the same algorithm described in the previous section, but
instead of applying the model we instead consult the user supplied oracle $y(p_i, p_j)$.
Of course, the oracle will not be able to supply a value for every pair - if so there would
be no need for a model! In the cases where the oracle does have some ground truth we save the
pairwise feature vector $\delta_{ij}$ and the value returned by the oracle, to generate
training and hold-out sets $\Delta_1$, $\Delta_2$. This algorithm is detailed in the Appendix.

We then train the model on the saved pairs, holding out some of the pairs for model evaluation.
Note: it is important to perform the holdout by splitting the set of points rather than pairs so that data from the same point does not
appear in both the training set and holdout set to avoid over-fitting. Our model is optimized for log-loss,
namely we seek to minimize the following quantity, with $\hat{y}_{ij}$ being the confidence score 
of our model (described in section~\ref{subsec:model}):
\begin{equation}
\vspace{-0.2cm}
    L = -\frac{1}{|\Delta|} \sum_{i,j \in \Delta} y_{i,j} \log{\hat{y_{ij}}} + (1 - y_{ij}) \log{(1 - \hat{y}_{ij})} .
\end{equation}

\subsection{Jointly learned node embeddings}
So far we have focused on the setting where we only use per-mode distance
scores as features, but this approach can be extended to also jointly learn
embeddings of the nodes themselves. In the extension we divide the
model into two components: the node embedding component and the similarity
learning component. The distances input to the similarity learning component are
augmented with per-dimension distances from the embeddings.
 By training
the embedding model with the similarity model, we can backpropagate
the similarity loss to the embedding layers obtaining
an embedding tailored for the task.

Augmenting the model in this way allows us to employ `two-tower' style networks\cite{bromley1994signature} to
learn similarity for features where that technique
works well (for example image data \cite{koch2015siamese}), but use pure distances for sparser features where
the embedding approach does not work as well. In some cases
we may even rely entirely on distances derived from the
embeddings (as is the case for some of the experiments in Section \ref{sec:academic_datasets}).
We note that the LSH family used above can still be applied in this setting
assuming that the embedding function is Lipschitz (following a similar
argument as in Section~\ref{subsec:lsh}).

\subsection{Model Structure}
\label{subsec:model}
The structure of the model used by \ourmethod\ can in general vary from
application to application.
In practice we have used \ourmethod\ with both neural networks and gradient
boosted decision tree models \cite{gbdt1,gbdt2}. Here we will describe 
the neural network approach in more detail.

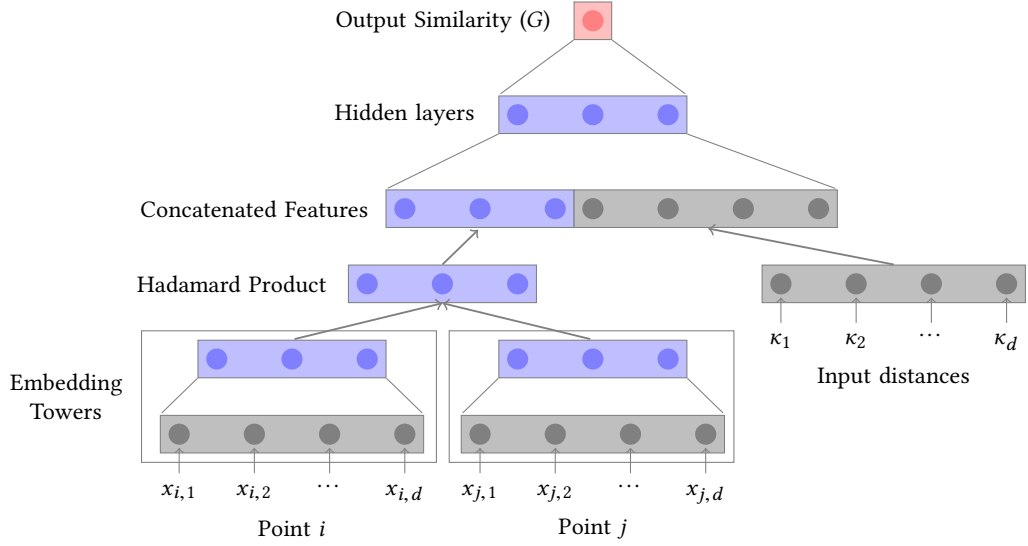
\begin{figure*}
\vspace{-0.4cm}
\centering
\def\gralelayersep{0.5cm}
\begin{tikzpicture}[shorten >=1pt,draw=black!50, node distance=\gralelayersep]

    \tikzstyle{every pin edge}=[<-,shorten <=1pt]
    \tikzstyle{neuron}=[circle,fill=black!25,minimum size=8pt,inner sep=0pt]
    \tikzstyle{pw neuron}=[neuron, fill=black!50];
    \tikzstyle{raw neuron}=[neuron, fill=black!50];
    \tikzstyle{output neuron}=[neuron, fill=red!50];
    \tikzstyle{hidden neuron}=[neuron, fill=blue!50];
    \tikzstyle{box}=[rectangle]
    \tikzstyle{annot} = [text width=8em, text centered]
    
    \node (XI-rect) [rectangle,draw,inner sep=0pt,fit={(0.75, -0.25) (4.25,0.25)},fill=black!25] {};
    \node[raw neuron, pin=below:$x_{i,1}$] (XI-1) at (1,0) {};
    \node[raw neuron, pin=below:$x_{i,2}$] (XI-2) at (2,0) {};
    \node[raw neuron, pin=below:\dots] (XI-3) at (3,0) {};
    \node[raw neuron, pin=below:$x_{i,d}$] (XI-4) at (4,0) {};
    
    \node (XJ-rect) [rectangle,draw,inner sep=0pt,fit={(4.75, -0.25) (8.25,0.25)},fill=black!25] {};
    \node[raw neuron, pin=below:$x_{j,1}$] (XJ-1) at (5,0) {};
    \node[raw neuron, pin=below:$x_{j,2}$] (XJ-2) at (6,0) {};
    \node[raw neuron, pin=below:\dots] (XJ-3) at (7,0) {};
    \node[raw neuron, pin=below:$x_{j,d}$] (XJ-4) at (8,0) {};

    \node (I-rect) [rectangle,draw,inner sep=0pt,fit={(8.75, 1.75) (12.25,2.25)},fill=black!25] {};
    \node[pw neuron, pin=below:$\kappa_1$] (K-1) at (9, 2) {};
    \node[pw neuron, pin=below:$\kappa_2$] (K-2) at (10, 2) {};
    \node[pw neuron, pin=below:\dots] (K-3) at (11, 2) {};
    \node[pw neuron, pin=below:$\kappa_d$] (K-4) at (12, 2) {};
    
    \node [xshift=0.5cm] (EHI-rect) [rectangle,draw,inner sep=0pt,fit={(0.75, 0.75) (3.25,1.25)},fill=blue!25] {};
    \foreach \name / \x in {1,...,3}
        \path[xshift=0.5cm]
           node[hidden neuron] (EHI-\name) at (\x, 1) {};
            
    \node [xshift=0.5cm] (EHJ-rect) [rectangle,draw,inner sep=0pt,fit={(4.75, 0.75) (7.25,1.25)},fill=blue!25] {};
    \foreach \name / \x in {1,...,3}
        \path[xshift=0.5cm]
            node[hidden neuron] (EHJ-\name) at (4 + \x, 1) {};
    
    \draw (0.5, -0.375) rectangle (4.42, 1.375); 
    \draw (4.59, -0.375) rectangle (8.375, 1.375); 
    
    \node (E-rect) [rectangle,draw,inner sep=0pt,fit={(3.25, 1.75) (5.75,2.25)},fill=blue!25] {};
     \foreach \name / \x in {1,...,3}
        \path
            node[hidden neuron] (E-\name) at (\x + 2.5, 2) {};    
    
    \node (combined_featuresL) [rectangle,draw,inner sep=0pt,fit={(3.75, 2.75) (6.25,3.25)},fill=blue!25] {};
    \foreach \name / \x in {1,...,3}
        \path[xshift=0.5cm]
            node[hidden neuron] (C-\name) at (\x + 2.5, 3) {};    
    \node (combined_featuresR) [rectangle,draw,inner sep=0pt,fit={(6.25, 2.75) (9.75,3.25)},fill=black!25] {};
    \foreach \name / \x in {1,...,4}
        \path[xshift=0.5cm]
            node[pw neuron] (D-\name) at (\x + 5, 3) {};            
    
    \node [xshift=0.5cm] (H-rect) [rectangle,draw,inner sep=0pt,fit={(4.75, 4.50) (7.25,4.0)},fill=blue!25] {};
    \foreach \name / \x in {1,...,3}
        \path[xshift=0.5cm]
            node[hidden neuron] (H-\name) at (\x + 4, 4.25) {};
            
    \node [xshift=0.5cm] (O-rect) [rectangle,draw,inner sep=0pt,fit={(5.75, 5.75) (6.25,5.25)},fill=red!25] {};
    \node[output neuron, xshift=0.5cm] (O) at (6, 5.5) {};
    
    \draw (EHI-rect.south east) -- (XI-rect.north east);
    \draw (EHI-rect.south west) -- (XI-rect.north west);
    
    \draw (EHJ-rect.south east) -- (XJ-rect.north east);
    \draw (EHJ-rect.south west) -- (XJ-rect.north west);
    
    \draw[<-, line width=0.25mm] (E-rect.south) -- (EHI-rect.north);
    \draw[<-, line width=0.25mm] (E-rect.south) -- (EHJ-rect.north);
    
    \draw[->, line width=0.25mm] (E-rect.north) -- (combined_featuresL.south);    
    \draw[->, line width=0.25mm] (I-rect.north) -- (combined_featuresR.south);

    \draw (H-rect.south west) -- (combined_featuresL.north west);
    \draw (H-rect.south east) -- (combined_featuresR.north east);

    
    \draw (O-rect.south east) -- (H-rect.north east);
    \draw (O-rect.south west) -- (H-rect.north west);

    \node[annot,left of=EHI-1, yshift=-0.5cm,text width=6em,node distance=2cm](tower1) {Embedding Towers};
    \node[annot,left of=E-1, node distance=1.8cm, text width=10em] (el) {Hadamard Product};
    \node[annot,left of=C-1, node distance=2.0cm, text width=10em] (el) {Concatenated Features};    
    \node[annot,left of=H-1, node distance=1.5cm] (hl) {Hidden layers};
    \node[annot,left of=O, node distance=2.0cm, text width=10em] (ol) {Output Similarity ($G$)};

    \node[annot,below of=K-2, xshift=0.5cm, node distance=1.25cm] (kl) {Input distances};
    \node[annot,below of=XI-2, xshift=0.5cm, node distance=1.25cm] (pointi) {Point $i$};
    \node[annot,below of=XJ-2, xshift=0.5cm, node distance=1.25cm] (pointj) {Point $j$};

\end{tikzpicture}
\vspace{-0.7cm}
\caption{The \ourmethod\ Neural Network model. The gray nodes are inputs, the blue are hidden layers, and red is the output. The network architecture combines
a standard two-tower model with natural distances in the input feature spaces. Weights are shared between
the two towers. The Hadamard product (pointwise multiplication) of the two towers is used to give us a pairwise embedding.
We treat this as an additional set of distance features to augment the input distance features $\kappa_1(x_i, x_j), \dots, \kappa_d(x_i, x_j)$.
This combined set acts as an input to the second part of the model which computes a single similarity score.}
\label{fig:grale_model}
\end{figure*}

The architecture of the neural net that we use is given in Figure \ref{fig:grale_model}.
The inputs to the neural net are split into two types: pairwise features which are functions
of the features of both nodes ($\delta_{ij}$ above), and singleton features - features of each node by itself.
Examples of pairwise features might include the Jaccard similarity of tokens associated with two nodes.
We use a standard `two-tower' architecture for the individual node features, sharing weights between each tower,
to build an embedding of each node. These embeddings are multiplied pointwise and concatenated with the vector of distance features $ \kappa_1(x_i, x_j), \dots, \kappa_d(x_i, x_j)$
to generate a combined pairwise set of features. We choose a pairwise multiplication here because it guarantees
that the model is symmetric in its arguments (i.e. $\hat{y}_{ij} = \hat{y}_{ji}$). Note that since the rest of
the model takes distances as inputs, it is symmetric by construction.

Note that here we have extended $G$ to be a function of not just the original distances but also of the input points
themselves. However, if we consider the embeddings learned as an augmentation of the data itself, then we are back to
the case of a pure function of distances, where we have also included coordinate-wise distances in the embedded space.
We can also argue that the locality sensitive hash functions for distances in the un-embedded versions of these spaces
cover as LSH functions in the embedded space due to the correlation between the two.

\subsection{Complexity}

Without the LSH process, the complexity of graph building on a set of $N$ points would be $\bigO{N^2}$ (since we compare
each point to every other). However instead, employing LSH we compute $S$ hashes per point.  Grouping points by hash value
is done by sorting, add a factor of $\bigO{SN \log{N}}$
To cap the amount of work done per bucket, we further subdivide these buckets into smaller subbuckets of maximum size $B$ 
($100$ is a typical value). Only then do we compare all points within the same subbucket, giving us a complexity of $\bigO{SNB}$ for this step.  Putting this together and doing a fixed amount of work per comparison, giving us a final complexity
of $\bigO{SN(B + \log{N})}$ work to score all of the edges.

\section{Evaluation on public datasets}
\label{sec:academic_datasets}
\subsection{Experimental Design}

There are many papers describing improved techniques for semi-supervised learning. We focus on  \cite{leman2018cikm} as a comparison
point since it also describes an approach to learning a task specific similarity measure, called PG-Learn. PG-Learn learns a set
of per dimension bandwidths for a generalized version of the RBF kernel
\begin{equation}
     K(x_i, x_j) = \exp\left(- \sum_{m} \frac{(x_{im} - x_{jm})^2}{\sigma_m}\right) .
\end{equation}
Rather than directly optimizing this as we do in this paper, PG-Learn sets out to optimize it as a similarity kernel for
the Local and Global Consistency (LGC) algorithm by Zhou et al \cite{LGCpaper}.

We compare on 2 of the same datasets used by \cite{leman2018cikm}.
USPS\footnote{\url{http://www.cs.huji.ac.il/~shais/datasets/ClassificationDatasets.html}} is a handwritten digit set, scanned from envelopes by the U.S. postal service and represented as  numeric pixel values.
MNIST\footnote{\url{http://yann.lecun.com/exdb/mnist/}} is another popular handwritten digit dataset, where the images have been size-normalized and
centered. Both tasks use a small sample of the labels to learn from, to more realistically simulate the SSL setting.
We list the size, dimension, and number of classes for these datasets in the appendix.  

To build a similarity model with \ourmethod, we select an oracle function that, for a pair of training nodes, 
the edge weight between them is 1 if they are in the same class and 0 otherwise.  Once trained, we use this
model to compute an edge weight for every pair of vertices. Since the number of nodes in these datasets is small
compared to the number of nodes in a typical \ourmethod\ application, we computed the edge weights for all
possible pairs of nodes; we did not use locality sensitive hashing.

We then use the graph as input to Equation \ref{ising_loss} and label nodes by computing a score for
each class as weighted nearest neighbors score, weighting by the log of the similarity scores. We choose
the class with the highest score as the predicted label. That is,
\begin{equation}
    \max_{c \in \mathcal{C}}\ \exp\left(\sum_{j \in N_c(i)} \log G_{ij}\right) ,
\end{equation}
where $N_c(i)$ is the set of nodes in the neighborhood of node $i$ that appear in our training set
with class $c$, and $G_{ij}$ is our model's similarity score for the pair.
Many of these other approaches also build a k-NN graph, with distance defined by a kernel:

In Table \ref{tab:academic_results}, we list the results of using \ourmethod\ for these classification tasks alongside the results from \cite{leman2018cikm},
which include PG-Learn, alongside some other approaches.

\begin{itemize}[leftmargin=0.25cm,itemindent=.25cm,labelwidth=\itemindent,labelsep=0cm,align=left,topsep=0pt,itemsep=-1ex,partopsep=1ex,parsep=1ex]

\item Grid search (GS): k-NN graph with RBF kernel where k and
bandwidth $\sigma$ are chosen via grid search, 
\item $Rand_d$
search (RS): k-NN with the generalized RBF kernel where k and different
bandwidths are randomly chosen, 
\item MinEnt: Minimum Entropy based tuning of the bandwidths
as proposed by Zhu et al \cite{zhuminent}, 
\item AEW: Adaptive Edge Weighting by Karasuyama et al. that
estimates the bandwidths through local linear reconstruction \cite{aew}.
\end{itemize}
For every result in this table, 10\% of the data was used for training.  For all results except \ourmethod, there was 15 minutes of automatic hyperparameter tuning.  For \ourmethod, we manually tested different edge weight models for each dataset.  For MNIST we used a two-tower DNN with two fully connected hidden layers. 
For USPS we used CNN towers.  

\begin{table}[t!]
  \vspace{-0.2cm}
  \begin{center}
    \begin{tabular}{l|c|c|c|c|c|c} 
      Dataset & \textbf{\ourmethod} & \textbf{PGLrn} & \textbf{MinEnt} &  \textbf{AEW} & \textbf{Grid} & \textbf{Rand$_d$}\\
      \hline
      USPS & 0.892 & 0.906 & $\mathbf{0.908}$ & 0.895 & 0.873 & 0.816 \\
      MNIST & $\mathbf{0.927}$ & 0.824 & 0.816 & 0.782 & 0.755 & 0.732\\
    \end{tabular}
  \end{center}
    \caption{Test Accuracy of Various Methods}
    \label{tab:academic_results}
    \vspace{-2em}
\end{table}

In Table \ref{tab:academic_results} we see that while \ourmethod\ does not always give the best performance
it produces results that are fairly similar to several other methods. In this case, the number of
labels is so small that we found it quite easy to overfit in similarity model training. It is likely
that in this case, all methods are learning roughly the same thing. This is especially true given how
competitive Grid is, where the similarity has a single free parameter.
On the other hand, MNIST provides quite a bit more to work with, so \ourmethod\ is able to learn
quite a bit and strongly outperforms the other methods.

\section{Case Study: \videosite\ Abuse Detection}

The comparisons in Section~\ref{sec:academic_datasets} give some sense for how \ourmethod\ 
performs; however, the datasets considered are much smaller than what \ourmethod\ was designed to
work with. 
They are also single mode problems, where the features belong to a
single dense feature space. 
We provide a case study here on how \ourmethod\ was applied to
abuse on \videosite, which is much more on the scale of problem that \ourmethod\ was
built to solve.

\videosite~is a video sharing platform with hundreds of millions of active users \cite{ytpress}.
Its popularity makes it a common target \cite{mccoy-show-me-the-money}
for organizations seeking to post to spam \cite{mccoy-click,samosseiko-partnerka,mccoy-pharmaleaks}.
Between April 2019 and June 2019, \videosite\ removed 4M channels for violating its
Community Guidelines \cite{yttransparency}.

\subsection{Experimental Setup}

We trained \ourmethod\ with ground truth given by Equation \ref{eq:youtube-grale-metric}, treating items
as either belong to the class of abusive items (removed for Community Guidelines violations)
or the class of safe (active, non-abusive) items. We also ignore the connections between safe
(non-abusive) pairs during training since these are not important for propagating abusive labels on the graph.
\begin{equation}
  \label{eq:youtube-grale-metric}
  y_{ij} =
  \begin{cases}
                                   1 & \text{if items $i$ and $j$ are abusive} \\
                                   0 & \text{otherwise} 
  \end{cases}
\end{equation}
Experiments were done using both the neural network and tree-based formulations of \ourmethod, but we
found better performance with the tree version.

\subsubsection{LSH Evaluation}
In our many applications of \ourmethod, we have observed that the choice of locality sensitive hash functions is critical to the overall
performance of the system.
There is some trade-off between $p$, the floor on the likelihood of producing
close pairs, and $q$, the ceiling of the likelihood of producing far pairs. Having too large $q$
results in graph building times that are too long to be useful, 
resulting in too much delay before items could be taken down. Alternatively, if $p$ is too small we miss
too many connections, hurting the overall recall of the entire system. 

Here we consider two different thresholds in evaluating our LSH scheme. The first ($r$ in Section~\ref{subsec:lsh})
is a threshold chosen such that a connection at that strength is strongly indicative of an abusive item. The 
second ('$cr$') is a more moderate threshold, chosen such that the connections are still interesting and useful when
combined with some other information in determining if an item is abusive. 
The final numbers we arrived at are given in Table \ref{tab:sketch_perf}.  In order to find the same
number of high weight edges using random sampling, we would need to evaluate $10.5x$ more edges than we
currently do (note this is assuming sampling without replacement, whereas LSH finds duplicates).

\begin{table}[t]
\vspace{-0.2cm}
\centering
\begin{tabular}{c|c|c}
    & \% strong ties (p) & \% weak ties (q)  \\
    \hline
    baseline (random pairs) & 0.0653\% & 77.4\% \\ 
    tuned LSH & 52.3\% & 22.7\% 
\end{tabular}
\caption{A comparison of the LSH function used for \videosite\ and a naive baseline. The parameters $p$ and $q$ are the same as defined
in Section~\ref{subsec:lsh}: \% of pairs returned by LSH with distance less than $r$ where $r$ is chosen to be a high precision decision threshold
of the model and \% of pairs returned by LSH with distance further than a moderate precision decision threshold, respectively. }
\label{tab:sketch_perf}
\end{table}

\subsection{Graph Analysis}
After finding a suitable hashing scheme and model for the dataset, we can materialize the graph.
Here we provide a quantitative analysis of the global properties of the graph.  In Section \ref{sec:visualization}, we visualize the graph for additional insights.

We set a threshold on edge weights to guarantee a very high level of precision. 
This means that many nodes end up without any neighbors in the graph. Table \ref{tab:proportions} shows that after
pruning 0-degree nodes (i.e. those without a very high confidence neighbor) the graph covers 36.96\%
of the abusive nodes, but only 2.31\% of the safe nodes.

\begin{figure}[t!]
\vspace{-0.4cm}
     \begin{subfigure}[b]{0.35\textwidth}
         \centering
        \includegraphics[width=\columnwidth]{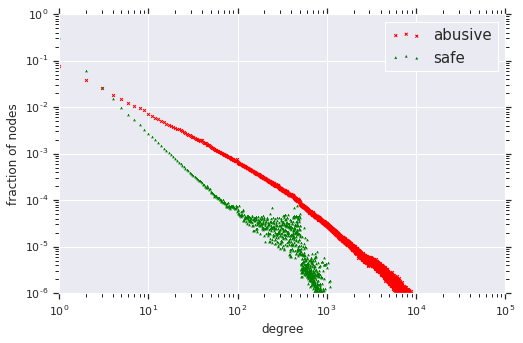}
        \caption{Node degree distribution in Grale YouTube.}
        \label{fig:degree-distribution}
\end{subfigure}
     \hfill
     \begin{subfigure}[b]{0.35\textwidth}
\centering
\includegraphics[width=\columnwidth]{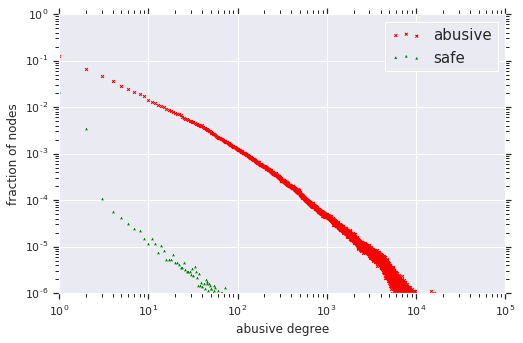}
\caption{Abusive-degree distribution in Grale YouTube.}
\label{fig:abusive-degree-distribution}     
     \end{subfigure}
    \caption{Degree distributions in the learned YouTube graph.  In \ref{fig:degree-distribution},
    the total degree distribution for nodes, broken down by abusive status. 
    In \ref{fig:abusive-degree-distribution}, the number of abusive neighbors for safe (green) and abusive (red) nodes.  We see that safe nodes have far fewer abusive neighbors on average.}
\end{figure}

\begin{table}[t]
    \vspace{-0.3cm}
    \centering
    \begin{tabular}{c|c|c|c}
        & degree 0 & degree >0 & high-precision degree >0 \\
        \hline
        safe (100\%) & 87.80\%	& 12.20\% &	2.32\% \\ 
        abusive (100\%) & 51.87\% &	48.13\% & 36.96\% \\
    \end{tabular}
    \caption{Percentage of items safe and abusive that have a node in the graph,
    considering only high-precision edges. Abusive nodes
    are 6-16x more likely to be connected in the graph.}
    \vspace{-0.5cm}
    \label{tab:proportions}
\end{table}

Figure~\ref{fig:degree-distribution} shows the distribution of node degrees, separated into safe and abusive items.
Abusive nodes generally have more neighbors than safe nodes.  However simply having related items doesn't
make an item abusive. Safe nodes may have other safe neighbors for legitimate reasons. 
So it is important to take into account more than just the blind structure of a node's neighborhood.

Similarly, Figure~\ref{fig:abusive-degree-distribution}  shows the degree of safe and
abusive items, but only when the neighbors are restricted to abusive nodes.
Safe nodes can have abusive neighbors if the edge evaluation model is imprecise or
the labels are noisy. However there is a clear drop-off in the frequency of 
high-degree safe nodes from Figure~\ref{fig:degree-distribution}.

\subsection{Classification Results}
\begin{figure}[t!]
\vspace{-0.5cm}
\centering
\includegraphics[width=0.35\textwidth]{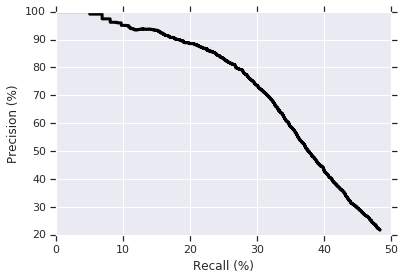}
\caption{Precision-recall curve of single nearest-neighbor classification on our graph. We selected
25\% of the abusive items as node seeds, and propagated the
label to neighboring nodes.}
\label{fig:1-hop-label-propagation}
\vspace{-0.5cm}
\end{figure}

Next, we performed a graph classification experiment using \ourmethod.
To show the utility of graphs designed by our method, we built and evaluated a simple single nearest neighbor classifier.\footnote{We note that in practice, a more complex system is employed, but the details of this are beyond the scope of this paper.}
We selected 25\% of the abusive nodes as seeds, and for every other node in the graph
we assigned a score equal to the maximum weight of the edges connecting
the node to a seed. 
Also, to simulate more real-world conditions we split our training
and evaluation sets by time, training the model on older data but evaluating
on newer data. This gives us a more accurate assessment of how the system
would function after deployment, since we will always be operating on data
generated after the model was trained.
The precision-recall curve of this classifier can
be observed in Figure \ref{fig:1-hop-label-propagation}.

We compared \ourmethod+Label Propagation, content classifiers,
and some heuristics against a subset of \videosite~abusive items.
Content classifiers were trained using the same training data to
predict abusive/safe. 
    
The results in Table \ref{tab:yt-abuse-2017} show that \ourmethod\ 
was responsible for half of the recall. While content classifiers
were given the first chance to detect an item (for technical reasons), Figure \ref{tab:yt-abuse-2017-age}
shows that content classifiers miss a significant part of recall when
faced with new trends which the classifiers have not been trained on.
With \ourmethod, the labels can be propagated to older items
as well without requiring new model retraining.

\begin{table}[t]
    \vspace{-0.4cm}
    \centering
    \begin{tabular}{c|c}
        Algorithm & \% of items \\
        \hline
        Content Classifiers & 47.7\% \\
        Heuristics & 5.3\% \\ 
        Total (Heuristics + Content Classifiers) & 53\% \\ 
        \hline
        \ourmethod+Label Propagation & 47.0\% (\textbf{+89\%})\\

    \end{tabular}
    \caption{Abuse identified from different techniques on a subset of \videosite\ abusive items.
    Adding Grale to the existing system increased recall by \textbf{89\%}.}
    \label{tab:yt-abuse-2017}
\end{table}

\begin{table}[t]
    \vspace{-0.4cm}
    \centering
    \begin{tabular}{c|c|c}
        Algorithm & New items & Old items \\
        \hline
        \ourmethod+Label Propagation & 25.8\% & 74.2\% \\
        Content Classifiers & 71.6\% & 28.4\% \\
        Heuristics & 33.7\% & 66.3\% \\ 
    \end{tabular}
    \caption{Breakdown by item age at the time of
    classification as abusive across various methods. `\ourmethod\ +
    Label Propagation is able to identify additional
    items initially missed by content classifiers.}
    \label{tab:yt-abuse-2017-age}
\vspace{-0.5cm}
\end{table}

\subsection{Graph Visualization}
\label{sec:visualization}

\begin{figure*}[t]
    \vspace{-0.3cm}
     \begin{subfigure}[b]{0.33\textwidth}
         \centering
        \includegraphics[height=1.5in,width=1.5in]{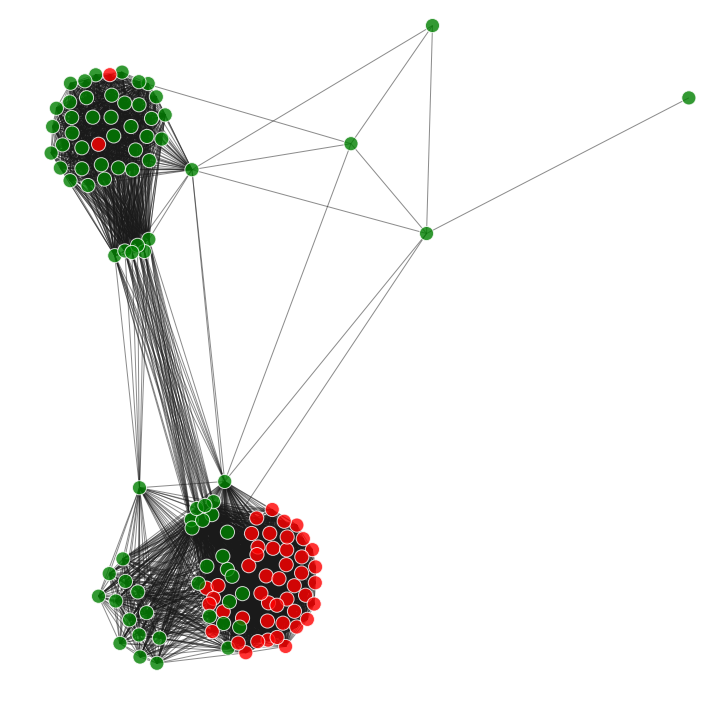}
        \caption{Intermixed dense clusters.}
        \label{fig:zen:intermixed}
    \end{subfigure}
     \hfill
     \begin{subfigure}[b]{0.33\textwidth}
        \centering
        \includegraphics[height=1.5in,width=1.5in]{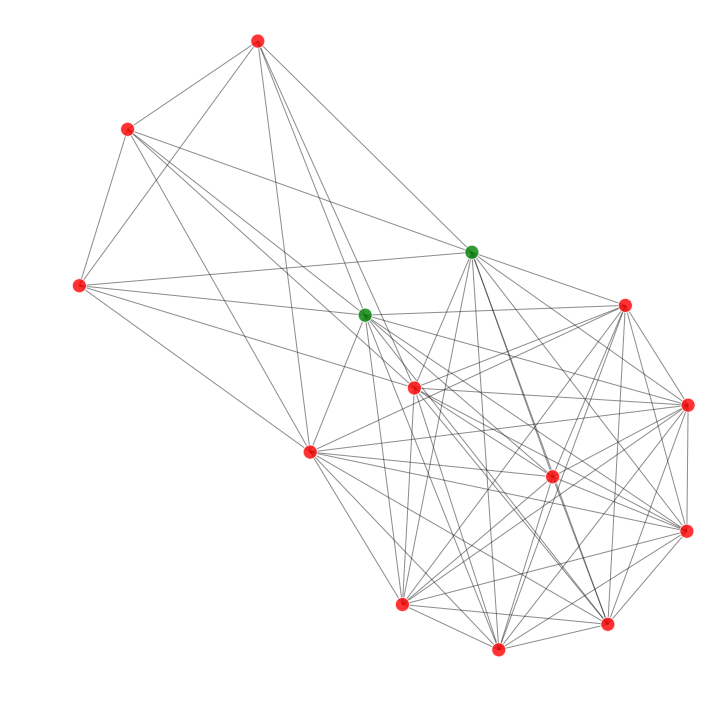}
        \caption{Sparse abusive cluster.}
     \end{subfigure}
     \begin{subfigure}[b]{0.33\textwidth}
        \centering
        \includegraphics[height=1.5in,width=1.5in]{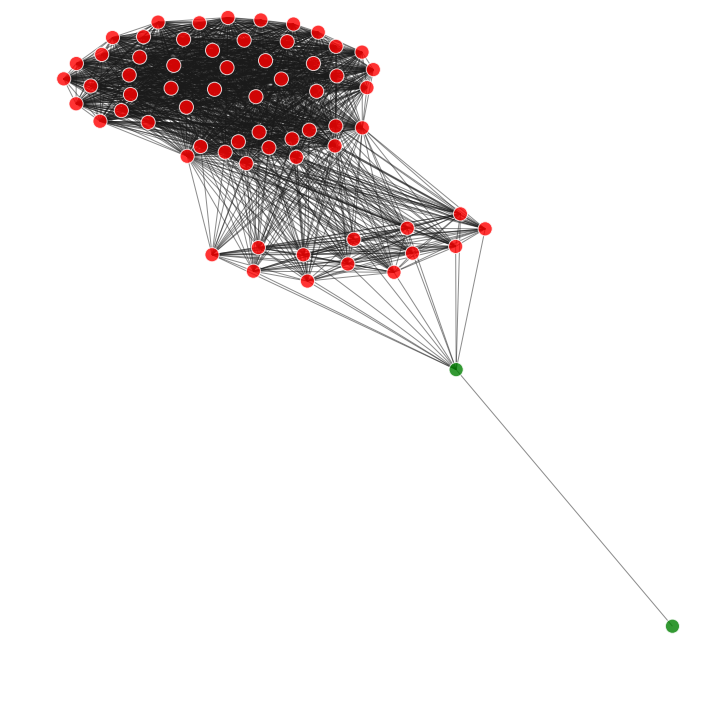}
        \caption{Dense abusive cluster.}
        \label{fig:zen:denseabusive}
     \end{subfigure}\\
     \begin{subfigure}[b]{0.33\textwidth}
         \centering
        \includegraphics[height=1.5in,width=1.5in]{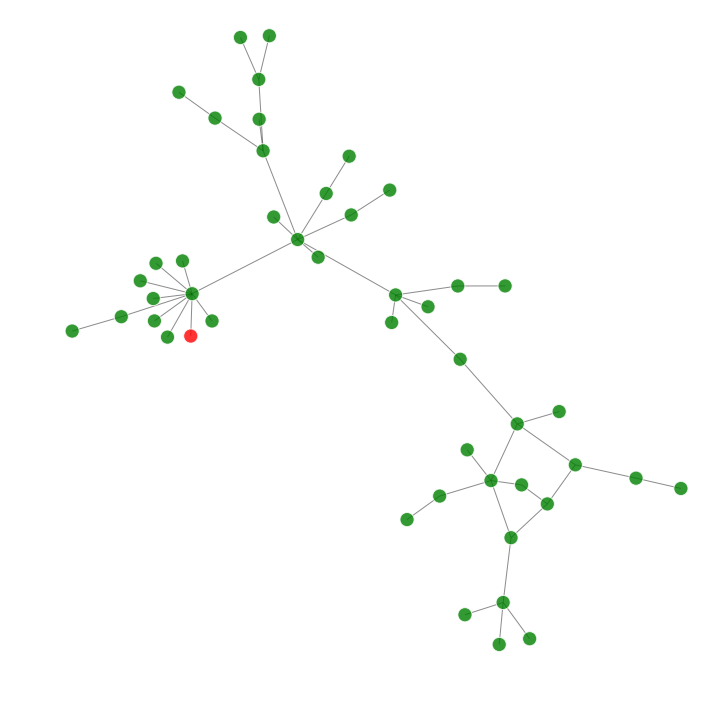}
        \caption{Sparse non-abusive subgraph.}
        \label{fig:zen:highdiameter}
    \end{subfigure}
     \hfill
     \begin{subfigure}[b]{0.33\textwidth}
        \centering
        \includegraphics[height=1.5in,width=1.5in]{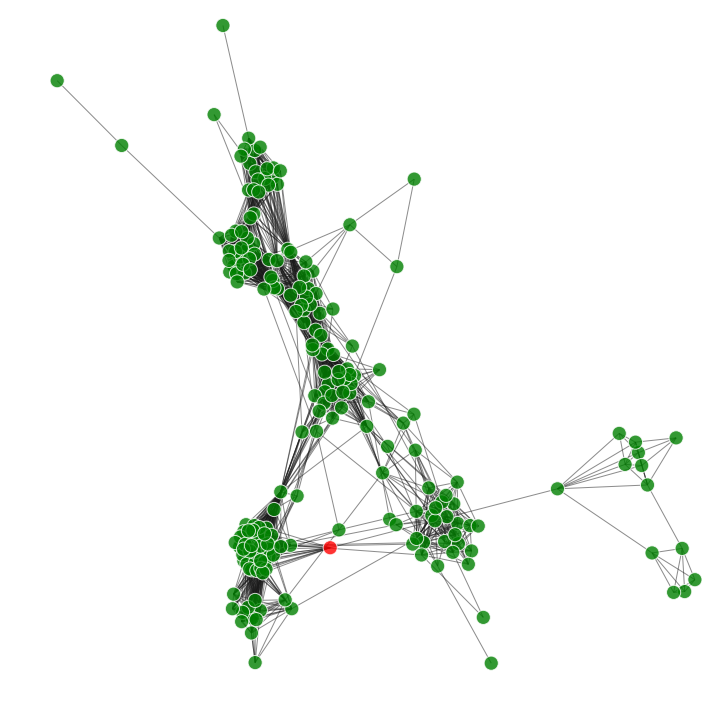}
        \caption{Dense non-abusive clusters.}
        \label{fig:zen:densenonabuseive}
     \end{subfigure}
     \begin{subfigure}[b]{0.33\textwidth}
        \centering
        \includegraphics[height=1.5in,width=1.5in]{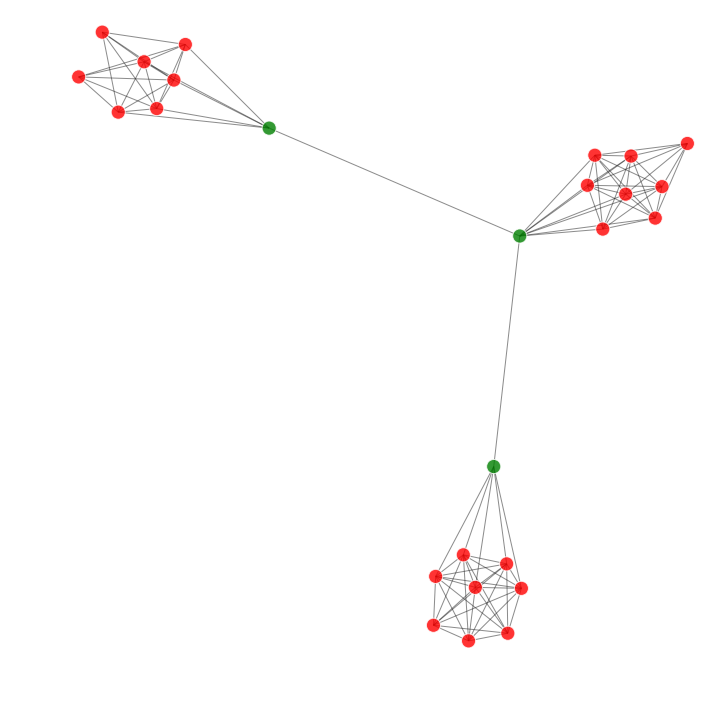}
        \caption{Small abusive clusters.}
     \end{subfigure}\\     
    \caption{Different subgraphs extracted from the YouTube case study.  Here we show a small sample of the rich patterns Grale is able to capture when applied to real data.  Colors correspond to the abuse status of the nodes.
    }
    \label{fig:zenpictures}
\end{figure*}

Finally, to give a sense of how well the scope and variety of structure present in graphs designed by \ourmethod, here we provide two kinds of visualizations of the local and global structure of
the \\ YouTube graph.

\subsubsection{Local structure}
Figure~\ref{fig:zenpictures} shows visualization of several subgraphs extracted from the YouTube graph.
The resulting visualizations illustrate the variety of different structure present in the data.
For instance in Figure~\ref{fig:zen:intermixed}, we see that clear community structure is visible -- there is one tight cluster of
abusive nodes (marked in red) and one tight cluster of non-abusive (in green). 
We note how this figure (along with Fig.~\ref{fig:zen:densenonabuseive}) illustrates the earlier point that existing in a dense neighborhood in the graph,
while potentially suspicious, is not necessarily an indicator of abuse.
Simultaneously, we can see that the resulting graph can contain components with relatively small or large diameter (Fig.~\ref{fig:zen:denseabusive} and Fig.~\ref{fig:zen:highdiameter}  respectively).

\subsubsection{Global structure}  
\begin{figure}[t]
\vspace{-0.2cm}
\centering
\includegraphics[width=0.8\columnwidth,trim=4cm 4cm 3cm 4cm,clip=true]{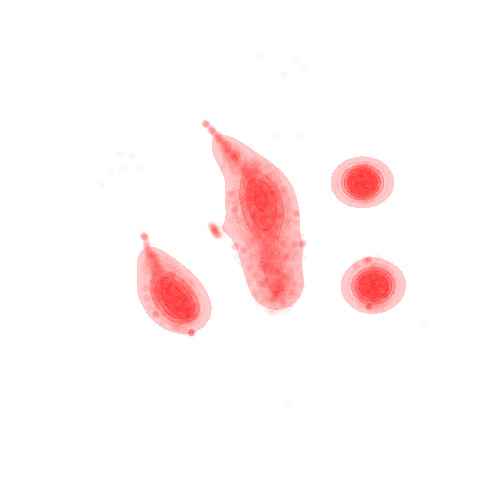}
\vspace{-1.0cm}
\caption{Visualization of global structure.  In order to visualize the global structure of the YouTube case study, we learned a random walk embedding \cite{perozzi2014deepwalk,Abu-El-Haija:2017:LER:3132847.3132959} of the highest weight edges in the graph.  Here we show a t-SNE projection of a sample of these embeddings.  Clusters correspond to known abusive behavior.}
\vspace{-1cm}
\label{fig:clamps-embedding}
\end{figure}

In order to visualize the global structure of the YouTube case study, we trained a random walk embedding \cite{perozzi2014deepwalk,Abu-El-Haija:2017:LER:3132847.3132959} on the YouTube graph learned through \ourmethod.  
Figure \ref{fig:clamps-embedding} shows a two-dimensional t-SNE \cite{maaten2008visualizing} projection of this embedding.  
We observe that, in addition to the rich local structure highlighted above, the Grale graph contains interesting global structure.
In order to understand the global structure of the graph better, we performed a manual inspection of dense regions of the embedding space, and found they corresponded to known abusive behavior.

\section{Related Work}
We begin by noting that the literature on label inference is vastly larger than that of graph construction.
In this section, we briefly discuss the relevant literature in graph construction, which we divide into unsupervised and supervised approaches.

\subsection{Unsupervised Graph Construction}

Perhaps the most natural form of graph construction is to take nearest neighbors in the underlying metric space.  
The two straightforward approaches to generate the neighborhood for each data item $v$ either connect it to the $k$-nearest data items, or to all neighbors within an $\epsilon$-ball.
While such \emph{proximity graphs} have some desirable properties, such methods have difficulty adapting to multimodal data.

For large datasets, the naive approach of comparing each point to every other point to find nearest
neighbors becomes intractably slow. Locality sensitive hashing is a classic technique for doing
approximate nearest neighbor search in metric spaces, speeding up the process significantly. The simplest
version of this algorithm would be to compare all pairs which share a hash value. However this still 
has worst-case complexity of $\bigO{N^2}$ ($N$ being the number of points in question). Zhang et al
\cite{zhang2013fast} propose a modification of this algorithm which reduces the complexity to 
$\bigO{N \log{N}}$ by sorting points by hash value and executing a brute force search within fixed
sized windows in this ordered list.

\subsection{Supervised Graph Construction}
A second branch of the literature seeks to use training labels to learn similarity scores between a set of nodes \cite{al2006link}.
Motivated by tasks like item recommendation \cite{chen2005link}, these methods aim to model the relationship between data items \cite{chen2018edges,chen2018tutorial,cui2018survey}, connecting new nodes or even removing incorrect edges from an existing graph \cite{Perozzi:2016:RGW:2939672.2939734}.
While \ourmethod\ could be thought of as a link prediction system, our motivations are substantially different.  
Instead of finding new connections between data items, we seek to accurately characterize the many possible interactions into the most useful form of similarity for a given problem domain.
Other work focuses on modeling the joint distribution between the data items and their label for better semi-supervised learning \cite{kingma2014semi}.
\citet{salimans2016improved} use a generative adversarial network (GAN) to model this distribution.
These approaches typically have good performance on small-scale SSL tasks, but are not scalable enough for the size or scale of real datasets, which may contain millions of distinct points.
Unlike this work, Grale is designed to be scalable and therefore utilizes a conditional distribution between the data and labels.

Perhaps the most relevant related work comes from 
\citet{leman2018cikm}, who propose a kernel for graph construction which reweights the columns of the feature matrix.
Unlike this approach, \ourmethod\ is capable of modeling arbitrary transformations over the feature space, operating on complex multi-modal features (like video and audio), and running on billions of data items.
Similarly, work on clustering attributed graphs have explored attributed subspaces \cite{Muller:2009:ECS:1687627.1687770} as a form of graph similarity, for example, using distance metric learning to learn edge weights \cite{Perozzi:2014:FCO:2623330.2623682}.

\section{Conclusion}

In this paper we have demonstrated \ourmethod, a system for learning
task specific similarity functions and building the associated graphs in
multi-modal settings with limited data.
 \ourmethod\ scales up to handling a graph
with billions of nodes thanks to the use of locality sensitive hashing,
greatly reducing the number of pairs that need to be considered.

We have also demonstrated the performance of \ourmethod\ in two ways. First, on smaller academic
datasets, we have shown that it is capable of producing competitive results in settings
with limited amounts of labeling. 
Second, we have demonstrated the capability and effectiveness of \ourmethod\ on a real \videosite\ dataset with hundreds of millions of data items.

\section*{Acknowledgements}

We thank Warren Schudy, Vahab Mirrokni, Natalia Ponomareva, Peter Englert, Filipe Miguel Gonçalves de Almeida, and Anton Tsitsulin.

\printbibliography

\newpage
\section*{Appendix}

\subsection*{Training Procedure}
Algorithm \ref{alg:training} shows how the sketching function is used in the train/test split.  

\begin{algorithm}[h]
\SetKwData{B}{Buckets}
\SetKwData{Points}{$\mathcal{X}$}
\SetKwData{GraleModel}{$\hat{y}$}
\SetKwData{TrainPoints}{$\mathcal{X}_{train}$}
\SetKwData{TestPoints}{$\mathcal{X}_{test}$}
\SetKwData{TrainPairs}{$\mathcal{X}_{train}$}
\SetKwData{TestPairs}{$\mathcal{X}_{test}$}
\SetKwData{TrainLabels}{$\mathcal{Y}_{train}$}
\SetKwData{TestLabels}{$\mathcal{Y}_{test}$}
\SetKwFunction{Oracle}{$\mathbf{Y}$}
\SetKwFunction{NNSketching}{NNSketching}
\SetKwInOut{Input}{Input}\SetKwInOut{Output}{Output}

\Input{A set of points $\Points$ and oracle function $\Oracle$}
\Output{A similarity model on approximating $\Oracle$ on $\Points$}
\BlankLine
Split $\Points$ into training and hold-out sets, $\TrainPoints, \TestPoints$\;
$\B = \NNSketching{\Points}$ \;
$\TrainPairs = \B \cap \TrainPoints$ \;
$\TestPairs = \B \cap \TestPoints$ \;
\tcc{Pairs which the oracle cannot decide are ignored during training}
$\TrainLabels =\Oracle{\TrainPairs}$ \;
$\TestLabels = \Oracle{\TestPairs}$ \;
Fit $\GraleModel$ on $(\TrainPairs, \TrainLabels)$\;
Evaluate model on $(\TestPairs, \TestLabels)$\;
\Return{$\GraleModel$}\;
\caption{Training procedure used for Grale, making using of the NNSketching function from Algorithm~\ref{alg:graphbuilder}.
Pairs which are not scorable by the oracle are omitted from its output. The model returned by this procedure
can then be used with Algorithm~\ref{alg:graphbuilder} to build a nearest neighbor graph.}
\label{alg:training}
\end{algorithm}

\subsection*{Dataset details}

Some details about the datasets used in Section \ref{sec:academic_datasets} are described here.

\begin{table}[h!]
  \vspace{-0.2cm}
  \begin{center}
    \begin{tabular}{l|c c c c} 
      Name & \# points & \# dimension & \# classes & Description\\
      \hline
      USPS & 1000 & 256 & 10 & Handwritten digits\\
      MNIST & 70000 & 784 & 10 & Handwritten digits\\
      \hline
    \end{tabular}
  \end{center}
    \caption{Datasets}
    \label{datasets}
  
\end{table}

\end{document}